\documentclass[letterpaper, 10 pt, conference]{ieeeconf}  

\IEEEoverridecommandlockouts                              

\usepackage[table,xcdraw]{xcolor}
\usepackage{multirow}
\usepackage{booktabs}
\usepackage{bm}
\usepackage{amsmath}
\usepackage{amssymb}
\usepackage{graphicx}
\usepackage{array}
\usepackage{mathtools}
\usepackage{hyperref}
\hypersetup{
    colorlinks=true,  
    linkcolor=black,  
    citecolor=black,  
    filecolor=black,  
    urlcolor=blue     
}
\usepackage{float}

\usepackage[nolist,nohyperlinks]{acronym}
\newacro{CFM}[CFM]{Conditional Flow Matching}
\newacro{GS}[GS]{Go Stanford}
\newacro{DDPM}[DDPM]{Denoising Diffusion Probabilistic Models}
\newacro{CNF}[CNF]{Continuous Normalizing Flows}
\newacro{FM}[FM]{Flow Matching}

\newacro{NOMAD}[NoMaD]{Navigation with Goal Masked Diffusion}
\newacro{ND}[NoMad-with-Depth]{}
\newacro{FD}[FlowNav-without-Depth]{}

\iffalse 
  \newcommand{\gode}[1]{\noindent}
  \newcommand{\nayak}[1]{\noindent}
  \newcommand{\oliveira}[1]{\noindent}
\else
  \newcommand{\gode}[1]{\textcolor{red}{\bf [SG: #1]}}
  \newcommand{\nayak}[1]{\textcolor{green}{\bf [AN: #1]}}
  \newcommand{\oliveira}[1]{\textcolor{magenta}{\bf [DO: #1]}}

\fi

\title{\LARGE \bf
FlowNav: Combining Flow Matching and Depth Priors\\for Efficient Navigation
}

\author{Samiran Gode*$^{1}$, Abhijeet Nayak*$^{1}$, Débora N.P. Oliveira*$^{1}$, Michael Krawez$^{1}$,\\Cordelia Schmid$^{2}$ and Wolfram Burgard$^{1}$
\thanks{* These authors contributed equally to the work}
\thanks{${1}$ Artificial Intelligence and Robotics Lab, Department of Computer Science and Artificial Intelligence, University of Technology Nuremberg, Germany. \{\tt\small\textit{firstname.lastname}\}\tt\small{@utn.de}}%
\thanks{${2}$ Inria, Ecole Normale Supérieure, CNRS, PSL Research University, France. \{\tt\small\textit{firstname.lastname}\}\tt\small{@inria.fr}}%
\thanks{$\dagger$ \url{https://utn-air.github.io/flownav.github.io/} }
}
\usepackage{changes} 

\usepackage{arydshln}
\usepackage{booktabs}

\begin{document}

\maketitle
\thispagestyle{empty}
\pagestyle{empty}

\begin{abstract}

Effective robot navigation in unseen environments is a challenging task that requires precise control actions at high frequencies. 
Recent advances have framed it as an image-goal-conditioned control problem, where the robot generates navigation actions using frontal RGB images. Current state-of-the-art methods in this area use diffusion policies to generate these control actions. Despite their promising results, these models are computationally expensive and suffer from weak perception. To address these limitations, we present FlowNav, a novel approach that uses a combination of \ac{CFM} and depth priors from off-the-shelf foundation models to learn action policies for robot navigation. FlowNav is significantly more accurate and faster at navigation and exploration than state-of-the-art methods. We validate our contributions using real robot experiments in multiple environments, demonstrating improved navigation reliability and accuracy. Code and trained models are publicly available$^\dagger$. 

\end{abstract}


\section{Introduction}
\label{sec:intro}

\bstctlcite{IEEEexample:BSTcontrol}

Autonomous navigation requires an agent to move efficiently and reliably from its current position to a goal position, while avoiding obstacles and following an optimal path~\cite{siegwart2011introduction}.
Classical approaches plan this optimally between the robot's current state and the goal state~\cite{Lat91Rob,choset05,lavalle06} by relying on a pre-built scene map~\cite{thrun2005probabilistic}.
Although these methods have been successful in various challenging scenarios~\cite{Bur99Exp,thrun06stanley,tranzatto2022cerberus}, they can fail to generalize to new environments~\cite{cadena2016past}.

\begin{figure}[!t]
    \centering
    \includegraphics[height=\linewidth,angle=90]{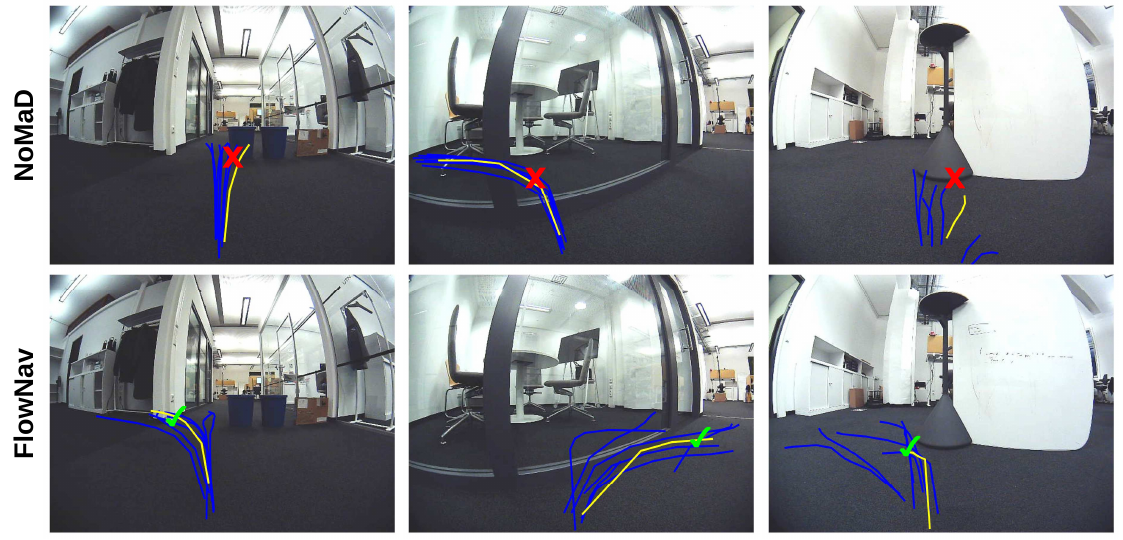}
    \caption{Comparison of sampled trajectories with goal-masked diffusion (NoMaD) and FlowNav (ours).
    Sampled trajectories are  in blue; yellow trajectories are executed.
    FlowNav uses a combination of CFM and depth priors to generate collision-free actions.}
    \label{fig:depth_comparison}
\end{figure}

Recent work~\cite{sridhar2024nomad,shah2023gnm,shah2023vint} addresses the domain generalization problem by training visual-navigation foundation models on large amounts of data~\cite{shah2023vint}.
An emerging trend in autonomous navigation~\cite{sridhar2024nomad,hiroselelan,doshi2024scaling} is to use diffusion policy~\cite{chi2023diffusion} for action generation.
Although diffusion ensures stable training~\cite{tong2023improving}, it has a high computational cost.

In contrast, Conditional Flow Matching (\ac{CFM}) has emerged as a more robust generalization of diffusion with faster inference times across various modalities~\cite{tong2023improving,lipman_flow_2022,albergo_building_2023,liu_rectified_2022,black2024pi0,chisari2024pointflowmatch}.
Recent work~\cite{esser2024scaling} also shows that flow matching excels at generating conditioned distributions, enabling the inclusion of priors that can help generate more informed action distributions.
Given these strengths of flow matching, we argue that it is well suited for robot navigation, especially in environments with obstacles. 

In this paper, we propose FlowNav, an image-conditioned action policy trained with \ac{CFM} to generate robot control actions. 
As depth perception and obstacle affordances are important for navigation, we introduce the use of embeddings from models for monocular depth estimation.
We hypothesize that models trained to estimate depth from images help the perception model discern depth differences between objects in the environment.
FlowNav significantly outperforms the diffusion-based baseline, as shown in Fig.~\ref{fig:depth_comparison}.
The main contributions of this work are as follows:
\begin{itemize}
    \item We present FlowNav, an efficient image-conditioned, cross-embodied, monocular-only action policy for autonomous navigation and exploration. We demonstrate significant improvements against state-of-the-art visual-navigation models through real-robot experiments.
    \item We propose the use of depth priors from foundation models for monocular depth estimation and report increased navigation and exploration performance over the baselines.
    \item We demonstrate that FlowNav trained without depth priors achieves shorter inference times, while matching the accuracy of state-of-the-art baselines.
    \item We release code, videos and the trained models.
\end{itemize}

\section{Related Work}
\label{sec:related_work}

\subsection{Classical Navigation}
Autonomous navigation has been a central research area in robotics, with various approaches addressing the challenge of guiding robots through complex environments~\cite{siegwart2011introduction}. Conventional methods that are based on the combination of  Simultaneous Localization and Mapping (SLAM)~\cite{thrun2005probabilistic,fox1999monte} and path planning techniques~\cite{latombe2012robot, lavalle1998rapidly} have demonstrated substantial robustness. 
However, these classical approaches often depend on hyperparameters that are environment specific, struggling to generalize across domains and environments.

\subsection{Learning-based Navigation}

Driven by advances in machine learning and the availability of large-scale data, there has been a shift toward learning-based methods for robot navigation~\cite{codevilla2018end}. 
End-to-end methods have focused on general goal-conditioned vision-based navigation~\cite{shah2022viking,shah2021ving,morin2023one}, directly mapping sensory inputs to actions that control the robot. 
For instance, RECON~\cite{shah2021rapid} provides long-horizon navigation without a map by comparing the embedding of the currently observed environment and a goal image.
Moreover, GNM~\cite{shah2023gnm} introduces zero-shot generalization to new robot embodiments by training on data collected across various robots in a normalized action space. 

A significant body of research has also included language instructions instead of goal images~\cite{hiroselelan}. 
SPOC~\cite{ehsani2024spoc} presents an autonomous agent trained with a shortest-path expert. 
Follow-up works~\cite{poliformer-zeng25a,hu2024flare} introduce a zero-shot policy for navigation using reinforcement learning fine-tuning. 
While learning from human-comprehensible commands is a promising research area, it is considered orthogonal work and falls outside our scope. 
We focus on solving the problem of visual navigation, where the robot should learn to predict actions based only on a frontal camera image. 

To improve performance, recent research has also experimented with the inclusion of other priors besides the raw output from the sensors. 
For instance, ImagineNav~\cite{zhao2025imaginenav} predicts the next viewpoint using a novel synthesis model, while ExAug~\cite{Hirose2023exaug} augments the input data by re-projecting the current image using different camera parameters. 
FlowNav also builds on the idea of incorporating additional environmental priors by using depth embeddings from off-the-shelf foundation models.

\subsection{Diffusion-based Navigation Models}
Advancements in generative modeling, particularly in diffusion models~\cite{ho2020denoising,rombach2022high}, have influenced robot navigation. 
For example, ViNT~\cite{shah2023vint} builds upon GNM~\cite{shah2023gnm} by using a diffusion policy to generate sub-goal images that are grounded in the action space. 
This idea of using sub-goal images enables kilometer-scale navigation~\cite{shah2022viking}. However, generating images from latent space samples can be computationally expensive. 
\ac{NOMAD}~\cite{sridhar2024nomad} uses diffusion policy~\cite{chi2023diffusion} to learn multi-modal actions rather than generating sub-goal images. 
NoMaD also introduces a novel masking idea to train a single network for goal-directed and exploratory navigation. 

While NoMaD has proven to be accurate in generating control actions, diffusion models are based on stochastic differential equations that require many passes through the network~\cite{tong2023improving}. 
This can lead to significant computational overhead, making them less suitable for real-time applications where efficiency is critical. 
In contrast, \ac{CFM}~\cite{lipman_flow_2022,albergo_building_2023,liu_rectified_2022,chisari2024pointflowmatch,black2024pi0}, which is a generalization of diffusion, offers a less expensive alternative. 
\ac{CFM} draws ``straighter'' paths to transform noise to the target distribution, thus reducing the number of iterative denoising steps. 
FlowNav demonstrates that \ac{CFM} is more robust and computationally efficient in  generating action policies than diffusion, which is particularly important in time-sensitive environments. 

\section{Overview}
\label{sec:overview}

\subsection{Problem Formulation}

Visual-navigation models try to solve the problem of goal-conditioned navigation, where a robot either explores its surroundings or navigates to a specified image goal. 
We define a visual navigation policy $\pi$, which takes as input a set of observations $\prescript{i}{}{O}$ and outputs actions $\prescript{i}{}{A}$, where $i$ denotes the current image index. 
We define $\prescript{i}{}{o}$ as the current observation image from the camera and $T_o$ as the past observation horizon.
The observation context includes all the images in the observation horizon $\prescript{i}{}{O} = \{\prescript{(i-T_o)}{}{o} : \prescript{i}{}{o}\}$.  The policy $\pi$ samples from the distribution $P(\prescript{i}{}{A}\mid \prescript{i}{}{O})$, where $\prescript{i}{}{A}$ is the set of predicted future actions at $i$ with a horizon of size $T_a$, such that $\prescript{i}{}{A}=\{\prescript{i}{}{a}:\prescript{i+T_a}{}{a} \}$, where $\prescript{i}{}{a}$ is a single waypoint.

\subsection{ Navigation with Goal Masked Diffusion (NoMaD)}
\ac{NOMAD}~\cite{sridhar2024nomad} uses diffusion policy to sample from the distribution \(P(\prescript{i}{}{A}\mid \prescript{i}{}{O})\), where \(A_i\) is a set of waypoints and \(O_i\) is the set of observation images. 
Diffusion policy uses \ac{DDPM}~\cite{chi2023diffusion}, which initially samples from a Gaussian distribution $q(x)$ and then progressively denoises the sample using a learned noise prediction model $\varepsilon_\theta$ to reach the desired distribution $p(x)$.
\ac{NOMAD}'s action model \( \varepsilon_\theta(\prescript{i}{}{A}_k, \prescript{i}{}{c}, k) \) with trainable parameters \( \theta \) predicts the noise added in the \( k \)-th step. Here $\prescript{i}{}{c}$ is the encoding of the observations and $\prescript{i}{}{A}_k$ is the action representation after $k$ steps of adding noise. 
The sample is denoised according to
\begin{equation}
\prescript{i}{}{A}_{(k-1)} = \alpha \left(\prescript{i}{}{A}_k - \gamma \varepsilon_\theta(\prescript{i}{}{A}_k,\prescript{i}{}{c}, k) + \mathcal{N}(0, \sigma^2 I) \right)
\label{eq:denoise_step}.
\end{equation}

To train the model \( \varepsilon_\theta(\prescript{i}{}{A}_k,\prescript{i}{}{c}, k) \) for this denoising process, \( k \) is randomly selected and used to get the corresponding noise \( \varepsilon_k \). 
Noise is then added to the sample \( \prescript{i}{}{A}_0 \) to obtain \( \prescript{i}{}{A}_k \). 
The noise model is trained by reducing 
\begin{equation}
\mathcal{L} = \text{MSE}\Big(\varepsilon_k, \varepsilon_\theta(\prescript{i}{}{A}_k,\prescript{i}{}{c}, k)\Big).
\label{eq:noise_prediction}
\end{equation}
In contrast, FlowNav uses \ac{CFM}~\cite{lipman_flow_2022,liu_rectified_2022} to map from the Gaussian distribution to the action distribution, which results in faster inference and more efficient training. 

\ac{NOMAD}'s perception model encodes $^iO$ and the goal image $^go$ using the EfficientNet architecture~\cite{tan2019efficientnet}, and passes it through a transformer model~\cite{vaswani2017attention} of four layers and four attention heads to generate the context embedding $^ic$. 
However, an accurate perception model also requires other semantic and geometric priors, which such a small model struggles to learn on its own. 
FlowNav addresses this problem by adding depth priors in the form of embeddings from off-the-shelf foundation models for monocular depth estimation.

\section{Methodology}
\label{sec:method}

\begin{figure*}[t]
    \centering
    \includegraphics[width=\textwidth]{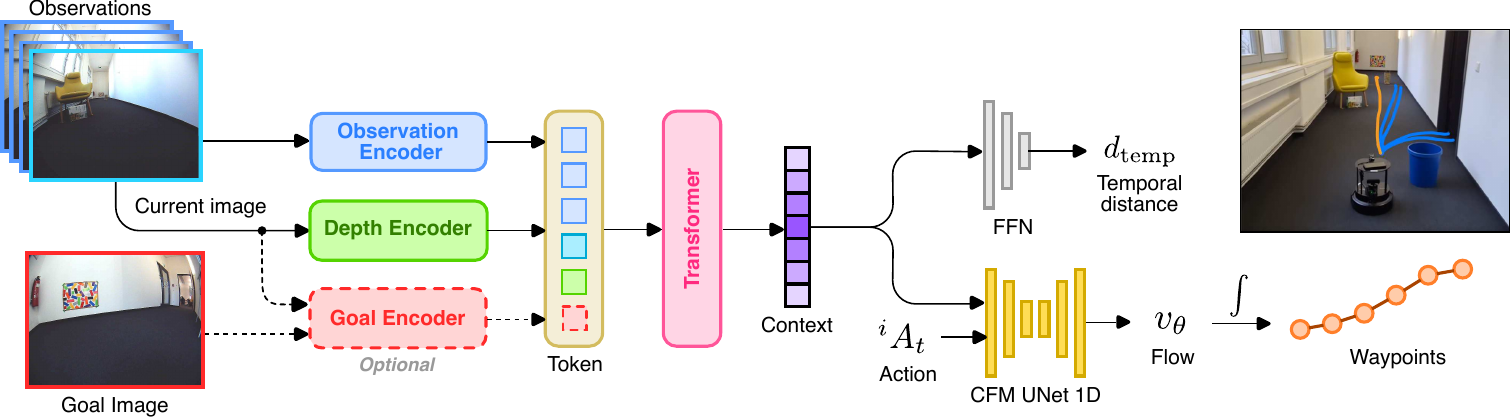}
    \caption{FlowNav Architecture:  
    We follow the architecture used in NoMaD~\cite{sridhar2024nomad}.
    In contrast to the diffusion policy used for action generation in \ac{NOMAD}, FlowNav uses Conditional Flow Matching (\ac{CFM}) which is more robust and efficient than diffusion.
    Additionally, we use depth priors from off-the-shelf foundation models for monocular depth estimation. 
    FlowNav first tokenizes the current and past observations, along with the goal image, using image encoders. 
    It also process the current observation with the pre-trained DinoV2 encoder from the Depth-Anything-V2~\cite{yang2025depth} model to generate the depth tokens.
    A transformer processes the observation, goal and depth tokens to create the observation context embedding $\prescript{i}{}{c}$. 
    We then use the context as input to the temporal distance prediction head $f_d$ and the velocity prediction network $v_\theta$. }
    \label{fig:arch}
\end{figure*}


In this section, we start with an introduction to the FlowNav architecture and show its advantages over the baseline. Then, we describe \ac{CFM}, a core component of FlowNav. 
Finally, we integrate these elements into the FlowNav policy, where we describe our action generation for navigation.

\subsection{FlowNav Architecture} \label{sec:flownav-arch}

Our architecture is inspired by \ac{NOMAD}~\cite{sridhar2024nomad} and is shown in Fig.~\ref{fig:arch}. 
We use an Efficient-NetB0~\cite{tan2019efficientnet} encoder, $f(I)$, to encode each image $ I\in \prescript{i}{}{O}$. 
We also concatenate the goal image $\prescript{g}{}{o}$ with the current image $\prescript{i}{}{o}$, which are then encoded using a second Efficient-NetB0 $g(\prescript{g}{}{o},\prescript{i}{}{o})$. 
Processing both the current observation and the goal image through the goal encoder helps the model perceive the difference between the current state and the required goal state~\cite{shah2023vint}.
During undirected exploration, the goal encoding is masked.

A novel addition to our architecture is the use of depth priors from off-the-shelf foundation models for monocular depth estimation.
We denote the feature encoder of this model as $h$.
For simplicity, we only encode the current observation $\prescript{i}{}{o}$.
Although the choice of model to use for our approach is arbitrary, owing to a vast amount of research on monocular depth estimation~\cite{Ranftl2022,yang2024depth,yang2025depth}, we use the pre-trained DinoV2~\cite{oquab2024dinov2} encoder from the Depth-Anything-V2 model~\cite{yang2025depth} because of the large amount of labeled and unlabeled data that the model was trained on.
We use the VIT-small version of the Depth-Anything-V2 model that generates a set of embeddings $h(\prescript{i}{}{o})$ $\in \mathbb{R}^{C \times D}$ with $D = 384$.
Further, we apply an adaptive average pooling along the channel dimension $C$ with a pooling dimension of 64 to summarize the embedding information.
The flattened embedding is then processed through an MLP to generate the depth tokens $\prescript{i}{}d$, which are used as inputs to the next block, the state encoder.

The state encoder, a transformer block $F$ puts together the observation, goal and depth tokens using a self-attention mechanism.
We choose a minimalistic transformer as used in~\cite{sridhar2024nomad} with four attention heads and four attention layers each, creating the observation context embeddings $\prescript{i}{}{c} = F\left(f\left(\prescript{i}{}{O}\right),g\left(\prescript{g}{}{o},\prescript{i}{}{o}\right),\prescript{i}{}{d}\right)$.

The context embedding $\prescript{i}{}{c}$ is processed by two separate prediction heads:
the temporal distance prediction head as in~\cite{sridhar2024nomad} and a velocity prediction head $v_\theta$ that is used for action generation.
The temporal distance prediction head $f_d(^ic)$ is an MLP that predicts the temporal distance between $^io$ and $^go$. 
The velocity prediction head predicts the flow used by \ac{CFM} with a 1D conditional U-Net. We explore the velocity prediction head in the following sections.

\subsection{Flow Matching Objective}\label{sec:flow-obj}
We define a smooth time varying vector field \( u : [0, 1] \times \mathbb{R}^d \to \mathbb{R}^d \)  as an Ordinary Differential Equation (ODE)
\begin{equation}
    dx = u_t(x) \, dt,
    \label{eq:displacement}
\end{equation}
where the solution is represented as \( \phi_t(x)\). At $t=0$,  \( \phi_0(x)= x_0 \), and moves based on velocity \( u_t \) to reach \( \phi_t(x) \) at time \( t \).
The objective is to learn the velocity \( u_t(x) \) that moves a point $x$ from an initial probability distribution \( q_0(x) \) to the desired distribution \( q_1(x) \). With this learned mapping, one can sample from a known distribution \( q_0(x) \) and reach the desired distribution \( q_1(x) \).

As explained in \cite{tong2023improving}, we assume that we know the probability path $p_t(x)$ and the velocity $u_t(x)$. If we define $v_\theta(x,t):[0,1]\times\mathbb{R}^d\rightarrow\mathbb{R}^d$ as a neural network with trainable parameters $\theta$, $u_t(x)$ can be regressed using the following \ac{FM} objective
\begin{equation}
\mathcal{L}_\text{FM}(\theta)=\mathbb{E}_{t \sim U(0,1), x \sim p_t(x)}\left \| v_{\theta}(x,t) - u_t(x) \right\|^2.
\label{eq:FM-loss}
\end{equation}
However, $u_t(x)$ is intractable. \ac{CFM} solves this problem by defining a marginal probability path $p_t(x\mid l)$ that varies according to some conditioning latent variable $l$, as in
\begin{align}
    p_t(x) &= \int p_t(x \mid l) q(l) \, dl,
    \label{p(x)_u(x)_defined_cfm}
\end{align}
where $q(l)$ is some distribution over the conditioning variable.
The \ac{CFM} objective is defined in \cite{tong2023improving,lipman_flow_2022} as 
\begin{equation}
    \mathcal{L}_\text{CFM}(\theta) := \mathbb{E}_{t, q(l), p_t(x \mid l)} \left[ \| v_{\theta}(t, x) - u_t(x \mid l) \|^2 \right],
    \label{eq: CFM_objective}
\end{equation}

With the conditioning variable $l$,  $p_t(x\mid l)$ and $u_t(x\mid l)$ are tractable, and it can be proven as shown in \cite{tong2023improving},\cite{lipman_flow_2022} that $\nabla_{\theta} \mathcal{L}_{\text{FM}}(\theta) = \nabla_{\theta} \mathcal{L}_{\text{CFM}}(\theta)$. 
Thus, by training $v_\theta(t,x)$ using the \ac{CFM} objective we regress $v_\theta(x,t)$ to $u_t(x)$.
At training time, we sample from \( x_1 \sim q_1(x) \) and \( x_0 \sim q_0(x) \). 
Like~\cite{tong2023improving}, for FlowNav we take $q(l) = q(x_0)q(x_1)$. 
The time \( t \) is sampled from \( U(0,1) \). We sample \( p_t(x\mid l) \) and \( u_t(x\mid l) \) using independent coupling~\cite{tong2023improving}
\begin{equation}
p_t(x \mid l) = \mathcal{N}\left(x \mid t x_1 + \left(1 - t\right)x_0, \sigma^2\right),
\label{eq:x_t_sample_with_sigma}
\end{equation}
\begin{equation}
u_t(x \mid l) = x_1 - x_0.
\label{eq:velocity_function}
\end{equation}
We use \(\sigma=0\) and thus \(x_t\) becomes
\begin{equation}
x_t =  tx_1 + (1 - t)x_0.
\label{eq:x_t_sample}
\end{equation}

\subsection{FlowNav Policy}

The FlowNav policy uses \ac{CFM} to sample from the distribution $P(\prescript{i}{}{A}\mid \prescript{i}{}{O})$ and follows the framework as in (\ref{eq:velocity_function}) and (\ref{eq:x_t_sample}). 
We define $^iA_t$ as the actions for image index $i$ transported from $^iA_0$ by the flow $u_t$ for time $t$.
The velocity prediction head predicts the velocity $v_\theta(t, ^iA_t,^ic)$ that moves the action $^iA_t$ at time $t$. 
The aim of the FlowNav policy is to transport $^iA_0$ sampled from the normal distribution at time $t = 0$ to the predicted robot waypoints $^iA_1$ at $t=1$. Given $^iA_0$ and $^iA_1$, we use $q(l)=q(\prescript{i}{}{A}_0)q(\prescript{i}{}{A}_1)$ as defined in Section \ref{sec:flow-obj}. 
According to (\ref{eq:velocity_function}), the flow in FlowNav transforms to 
\begin{equation}
    u_t(x\mid l) = \prescript{i}{}{A}_1 - \prescript{i}{}{A}_0.
\end{equation}

According to (\ref{eq:x_t_sample}), we can then sample an action $\prescript{i}{}{A}_t$ at some $t$ using 
\begin{equation}
    \prescript{i}{}{A}_t = t \prescript{i}{}{A}_1 + (1-t)\prescript{i}{}{A}_0,
\end{equation}
where $t$ is the time sampled from the distribution $U\sim(0,1)$.
The flow matching objective in (\ref{eq: CFM_objective}) transforms into
\begin{align}
\mathcal{L}_{\text{CFM}}(\theta) &= 
        \mathbb{E}_{p(\prescript{i}{}{A}_t \mid \prescript{i}{}{O}, \prescript{i}{}{A}_0, \prescript{i}{}{A}_1), q(l),t} \\
        &\quad \left[ \| v_{\theta}(t,\prescript{i}{}{A}_t, \prescript{i}{}{c}) - u_t(\prescript{i}{}{A}_t \mid \prescript{i}{}{A}_0, \prescript{i}{}{A}_1) \|^2 \right].\nonumber
\label{eq:flownav}
\end{align}
The temporal distance objective is defined as
\begin{equation}
\mathcal{L}_{\text{dist}}(\theta) = 
        \lambda \cdot \text{MSE}\Big(\text{dist}_\text{temp}(\prescript{i}{}{o}, \prescript{g}{}{o}), f_d(\prescript{i}{}{c})\Big).
\end{equation} 
The velocity prediction head is trained in conjunction with the temporal distance prediction head in an end-to-end supervised manner.
Our final loss therefore is
\begin{equation}
    \mathcal{L}(\theta) = \mathcal{L}_{\text{CFM}}(\theta) + \mathcal{L}_{\text{dist}}(\theta).
\end{equation}

During inference, assuming an image index $j$, we sample $\prescript{j}{}{A}_0$ from a normal distribution with an action horizon $T_a$. 
The sampled trajectories are defined in a 2D plane with shape $(T_a,2)$. 
The sample $\prescript{j}{}{A}_0$ is then transformed into the set of real robot actions $\prescript{j}{}{A}_1$ conditioned on the context $\prescript{j}{}{c}$ using the velocity prediction $v_\theta(t,\prescript{j}{}{A}_t,\prescript{j}{}{c})$.  
Although various solvers such as RK-4 and Dopri-5 are available for integration, we find that the first-order Euler update yields the best trade-off between prediction accuracy and inference time. 
The Euler update is defined as
\begin{equation}
    \prescript{j}{}{A}_{t+\delta} = \prescript{j}{}{A}_t + \delta v_\theta(t,\prescript{j}{}{A}_t,\prescript{j}{}{c}).
    \label{eq:Euler_step}
\end{equation}
  
The predicted waypoints are un-normalized using a robot-specific controller as in \cite{sridhar2024nomad}.
We define the number of $\delta$ steps applied to move $\prescript{i}{}{A}_{0}$ to $\prescript{i}{}{A}_{1}$ as the number of inference steps and is denoted as $\mathcal{K}$.

As in NoMaD~\cite{sridhar2024nomad}, FlowNav requires a precollected topological map of the environment for goal-directed navigation. 
Each node in the map is represented by an image. 
The temporal distance prediction head predicts the temporal distance between the current observation and the nodes in the topological map.
The node with the least temporal distance is used as the goal image for action prediction.

For task-agnostic exploration, we mask out the goal encodings as described in \cite{sridhar2024nomad}, as the action prediction head does not require a goal image.  
As a result, FlowNav learns to output multi-modal action distributions if there are multiple exploratory paths available in the environment.
The exploration mode is also used to store a topological map of the environment and can be used in conjunction with the navigation mode to enable long-horizon navigation.

\section{Experiments}
\label{sec:experiments}

\subsection{Implementation Details}
Our implementation builds on top of \ac{NOMAD}, where we update the training and evaluation pipelines to use \ac{CFM} instead of diffusion policy. 
For our CFM implementation we use the TorchCFM~\cite{tong2023improving, tong2023simulation} library. 
We train four different models on the same data splits: \ac{NOMAD}, which serves as our baseline; a depth-conditioned version of \ac{NOMAD}, referred to as \ac{NOMAD}-with-depth; FlowNav-without-depth, which utilizes \ac{CFM} without depth priors; and finally, FlowNav which incorporates both \ac{CFM} and depth priors.
We use code provided by the authors of the baseline and train their model according to their training strategy.

For action prediction, we use a prediction horizon of $T_a = 8$ waypoints with respect to the local robot frame. 
To maintain consistency, we define $T_o = 3$ which is the same as our baseline.
We train each model for 30 epochs on an NVIDIA H100 GPU with a batch size of 512. 
We use the AdamW~\cite{loshchilov2019decoupled} optimizer with an initial learning rate of 1e-4. 
The models were initially trained using a cosine scheduler with warmup~\cite{loshchilov2022sgdr}  as in \cite{sridhar2024nomad}.
However, we found that this makes the flow matching training unstable. 
We believe this occurs because \ac{CFM} trains faster than diffusion, causing the action head to overfit to the context from the perception head before it has fully converged. 
We ablated with multiple other scheduler variants and found that half-cosine with warmup and  1cycle~\cite{smith2019super} perform equally well. 
For FlowNav-without-depth, we empirically found that training the model for 17 epochs provide best performance. For FlowNav, we found that the additional depth information regularizes the model, and as a result, trains in 30 epochs.

\subsection{Datasets}
We use a combination of 5 open-source datasets to train all models: Go Stanford~\cite{hirose2018gonet}, RECON~\cite{shah2021rapid}, Tartan Drive~\cite{triest2022tartandrive}, SACSoN~\cite{hirose2023sacson} and SCAND~\cite{karnan2022socially}.
These datasets were selected because of the wide variety of indoor and outdoor cross-embodied data.
This helps the model generalize to new environments.
Each of these datasets contains a set of robot trajectories along with the captured frontal RGB images. 
We randomly select segments of trajectories as the observation context and a future waypoint as the goal location. 
Following \cite{sridhar2024nomad,shah2023gnm}, these segments are normalized, so that the model can be trained on diverse, cross-embodied datasets.
To make it a fair comparison against the baseline, we train it from scratch on the same data.

\subsection{Deployment}
\begin{figure}[t]
    \centering
    \includegraphics[width=\linewidth,scale=0.5]{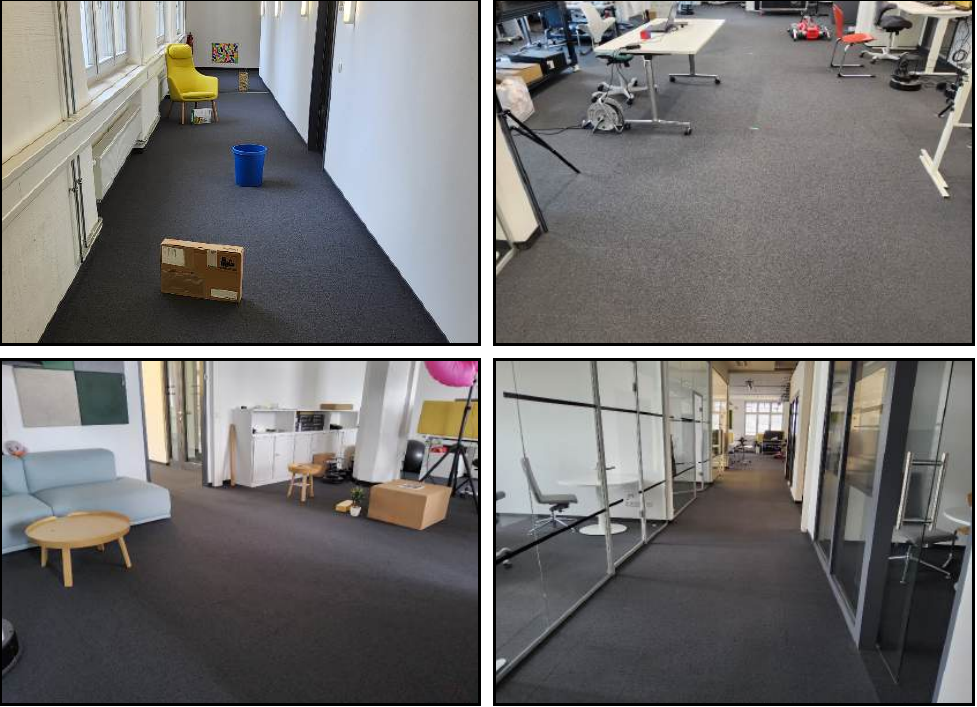}
    \caption{Examples of environments. We include easy (boxes and sofas) and hard obstacles (tables and chairs). Our test environment also involves navigating through glass walls and cluttered environments.}
    \label{fig:env_image}
\end{figure}

For real-robot experiments, all trained models are deployed on a TurtleBot4.
We use ROS 2 Humble in Discovery Server mode to minimize latency between action prediction and execution.
Following our baseline~\cite{sridhar2024nomad}, we use a wide-angle camera capturing images at 15 Hz.
The inference runs on an NVIDIA RTX A500 laptop GPU.
We used 4 different environments with difficulties ranging from easy to hard to evaluate all models. 
The test scenarios include soft and tight turns, compact obstacles (e.g. boxes), and hard-to-detect obstructions (e.g. chairs, table legs and glass walls). Some of our test scenarios are illustrated in Fig.~\ref{fig:env_image}.

\subsection{Experimental Results}

\begin{figure*}[t]
    \centering
    \includegraphics[height=\linewidth,angle=90]{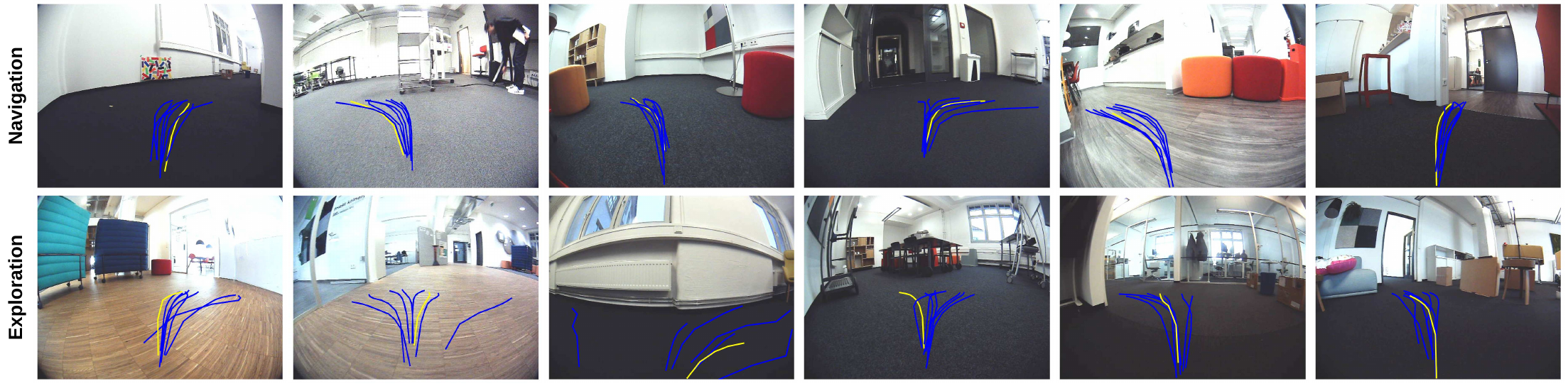} 
    \caption{Visualizing rollouts of FlowNav in challenging environments. 
    We sample 8 actions which are shown in blue. A randomly selected action (yellow) is executed on the robot. Note that providing the goal image clusters the trajectories together, while in exploration, the trajectories are multi-modal. }
    \label{fig:flow-examples}
\end{figure*}

We report results of all models for the goal-directed navigation and in the exploratory mode.
For goal-directed navigation, we report the goal-reaching success rates and the number of minor collisions per run.
We consider collisions that the robot cannot recover from as major collisions, for instance, getting stuck for more than 10 seconds or frontal collisions.
For the exploration experiments, the robot follows a closed path. 
In this case, we record the number of collisions that occur when the robot explores the path.

\subsubsection{Baseline Comparison}

Since NoMaD outperforms~\cite{sridhar2024nomad} the previous models GNM~\cite{shah2023gnm} and ViNT~\cite{shah2023vint}, we compare to NoMaD as the sole baseline.
Table \ref{table:nav_and_explore} compares all trained models on navigation and exploration experiments using $\mathcal{K} = 10$ inference steps for action generation. 
The results show that FlowNav outperforms \ac{NOMAD} and NoMaD-with-depth in all metrics.
We also show that FlowNav-without-depth has a success rate comparable to \ac{NOMAD}.
Moreover, NoMaD-with-depth and FlowNav have far fewer collisions than the plain models.
We observe a similar trend in the exploration experiments, where FlowNav model has just a single collision per run, which is a significant improvement over the plain models.
This reinstates our hypothesis that including depth priors during model training helps the model learn better action policies.

\subsubsection{Qualitative Results}

Fig.~\ref{fig:flow-examples} illustrates trajectories generated for navigation and exploration by  FlowNav in various environments.
The images show a set of generated actions (in blue and yellow).
The robot executes the action in yellow, which is selected at random.
FlowNav accurately learns contextually aware actions for exploration and navigation.

\begin{figure}[!t]
    \centering
    \includegraphics[height=\linewidth,angle=90]{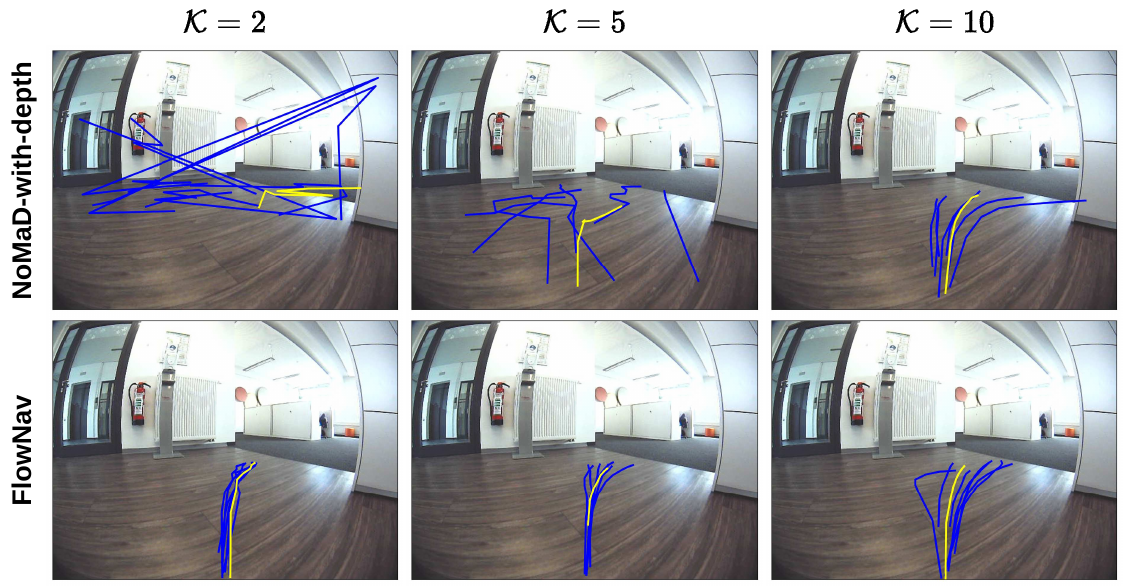}
    \caption{Comparison of generated trajectories with NoMaD-with-depth and FlowNav for $\mathcal{K}=2, 5$ and 10 inference steps. FlowNav produces valid trajectories with fewer steps, showing the effectiveness of flow matching.}
    \label{fig:k_variations}
\end{figure}

\subsubsection{Training Data Ablation}

Since we train all our models on a subset of the original \ac{NOMAD} training data, we investigate whether this limited dataset is the reason for a higher number of collisions for the plain models.
To do this, we use the pre-trained \ac{NOMAD} model in our exploration experiments.
With the pre-trained \ac{NOMAD} weights, we observe 3 collisions per run.
This is fewer than the 5 collisions per run that we see in \ac{NOMAD} model trained on our data subset, but still higher than the 1 collision per run we observe with FlowNav, and 1.5 with NoMaD-with-depth.
It is highly likely that using a larger training dataset helps the model generalize better to new environments.

\begin{table}[b]
    \centering
    \caption{Success rate and no. of collisions per run for navigation and exploration modes}
    \renewcommand{\arraystretch}{1.1}\setlength{\tabcolsep}{3.5pt}
    \begin{tabular}{l|c|c|c}
        \toprule
        \multirow{2}{*}{\textbf{Method}} & \multicolumn{2}{c|}{\textbf{Navigation}} & \textbf{Exploration} \\
         & Success $\uparrow$ & Coll. / run $\downarrow$ & Coll. / run $\downarrow$ \\ \hline
        \ac{NOMAD}~\cite{sridhar2024nomad} & 0.8 & 0.7 & 5.0 \\
        NoMad-with-depth& 0.9 & 0.6 & 1.5 \\ \hdashline
        \rowcolor[HTML]{dcfce7} FlowNav-without-depth (ours) & \textbf{1.0} & 0.7 & 4.5 \\
        \rowcolor[HTML]{dcfce7} FlowNav (ours) & \textbf{1.0} & \textbf{0.2} & \textbf{1.0} \\
        \bottomrule
    \end{tabular}
    \label{table:nav_and_explore}
\end{table}

\subsubsection{Inference Steps Ablation} 

To examine how the number of inference steps $\mathcal{K}$ impacts model performance, we conduct experiments in simplified environments with soft turns and no obstacles. 
Since we have already demonstrated that depth-conditioned models perform better, we limit our comparison to these models. 
We evaluate navigation and exploration tasks while varying $\mathcal{K}$.
For navigation, we again track success rates and the number of minor collisions per run, as in the previous experiment. 
For exploration, as we use environments without obstacles, we only track if exploration was successful (no collisions).
The results of these experiments are presented in Tables~\ref{table:nav_k_variation} and~\ref{table:explore_k_variation}.  
We observe that NoMaD-with-depth performs poorly with a small $\mathcal{K}$ but achieves a high success and a low collision rate at $\mathcal{K} = 10$. 
In contrast, FlowNav performs well even at \(\mathcal{K} = 2\) with a $75\%$ success rate, and continues to improve as $\mathcal{K}$ increases. 
We observe a similar trend in the exploration experiments, where FlowNav performs well at \(\mathcal{K} = 2\), while NoMaD-with-depth struggles to explore the environment due to frequent collisions.
These findings highlight that flow matching policies remain effective with fewer inference steps, thus reducing trajectory generation time. Fig.~\ref{fig:k_variations} visually represents our findings.

\begin{table}[t]
    \centering
    \caption{Success rate and collisions per run against no. of inference steps for navigation mode}
    \setlength{\tabcolsep}{5pt}
    \renewcommand{\arraystretch}{1.1}
    \begin{tabular}{l|cc|cc|cc}
        \toprule        
        \multirow{2}{*}{\textbf{Method}} & \multicolumn{2}{c|}{$\mathcal{K}$ = 2} & \multicolumn{2}{c|}{$\mathcal{K}$ = 5} & \multicolumn{2}{c}{$\mathcal{K}$ = 10} \\
        & SR $\uparrow$ & \# C $\downarrow$ & SR $\uparrow$ & \# C $\downarrow$ & SR $\uparrow$ & \# C $\downarrow$\\
        \hline
        NoMad-with-depth& 0.0 & 1.0 & 0.25 & 1.0 & \textbf{1.0} & \textbf{0.0} \\ \hdashline
        \rowcolor[HTML]{dcfce7} FlowNav (ours) & \textbf{0.75} & \textbf{0.25} & \textbf{1.0} & \textbf{0.0} & \textbf{1.0} & \textbf{0.0} \\
        \bottomrule
    \end{tabular}
    \label{table:nav_k_variation}
\end{table}

\begin{table}[t]
    \centering
    \caption{Success rate against the no. of inference steps in exploration mode} 
    \setlength{\tabcolsep}{12pt} 
    \renewcommand{\arraystretch}{1.1}
    \begin{tabular}{l|c|c|c}
        \toprule
        \multirow{2}{*}{\textbf{Method}} & $\mathcal{K} = 2$ & $\mathcal{K} = 5$ & $\mathcal{K} = 10$\\
         & SR $\uparrow$& SR $\uparrow$& SR $\uparrow$\\
        \hline
        NoMad-with-depth & 0 & 0.75 & \textbf{1.0} \\\hdashline
        \rowcolor[HTML]{dcfce7}FlowNav (ours) & \textbf{1.0} & \textbf{1.0} & \textbf{1.0} \\
        \bottomrule
    \end{tabular}
    \label{table:explore_k_variation}    
\end{table}

\subsubsection{Inference Time} 

Table~\ref{table:inference_times} shows action runtime with different numbers of inference steps $\mathcal{K}$. 
We show in Tables~\ref{table:nav_k_variation}, \ref{table:explore_k_variation} and Fig.~\ref{fig:k_variations} that FlowNav at $\mathcal{K}=2$ is comparable to NoMaD-with-depth at $\mathcal{K}=10$.
Thus, even though FlowNav has comparable runtime to NoMaD for the same $\mathcal{K}$, it requires less steps to beat the performance of NoMaD.

\begin{table}[t]
\centering
\caption{Inference times (in ms) for different no. of steps}
\renewcommand{\arraystretch}{1.1}\setlength{\tabcolsep}{16.5pt}
\begin{tabular}{l|c|c|c}
\toprule
\multicolumn{1}{c|}{$\mathcal{K}$} & 2 & 5 & 10 \\
\hline
\ac{NOMAD}~\cite{sridhar2024nomad} & 17.2 & 26.7 & 44.8 \\ \hdashline
\rowcolor[HTML]{dcfce7} FlowNav (ours) & 19 & 26.4 & 48.6 \\
\bottomrule
\end{tabular}
\label{table:inference_times}
\end{table}


\section{Conclusions}
\label{sec:conclusion}

\hyphenation{FlowNav}

In this paper, we present FlowNav as a novel and image-conditioned action policy that is trained with Conditional Flow Matching (\ac{CFM}) and additionally integrates depth priors. The improved efficiency obtained through the usage of \ac{CFM} and the addition of the depth priors leads to a better performance and faster runtimes in both navigation and exploration compared to the state-of-the-art approach NoMaD.
We validate this through extensive real-robot experiments.
Our results show that FlowNav consistently outperforms prior approaches, setting a new standard for robust and efficient navigation.
For future work, we would like to focus on using off-policy reinforcement learning to learn an optimal trajectory for goal-based exploration. We plan to substitute the topological map with implicit representations learned with the same transformer backbone.

\section*{Acknowledgments}
 The authors gratefully acknowledge the scientific support and HPC resources provided by the Erlangen National High Performance Computing Center (NHR@FAU) of the Friedrich-Alexander-Universität Erlangen-Nürnberg (FAU) under the BayernKI project v106be. BayernKI funding is provided by Bavarian state authorities. We also acknowledge the support by the Körber European Science Prize.


\bibliographystyle{IEEEtran}
\bibliography{root}

\end{document}